# 多视角特征融合的网络异常流量检测

宋 昊[1,2] 傅文涛[1,2] 陈烜泽[1,2] 金程祥[1,2] 周嘉俊[1,2,*] 俞山青[1,2] 宣 琦[1,2]

（1. 浙江工业大学 网络空间安全研究院，杭州 310023；2. 杭州市滨江区浙工大人工智能创新研究院，杭州 310056）

**摘 要** 传统的异常流量检测方法基于单一视角分析，在处理复杂攻击和加密通信时具有明显的局限性。提出一种多视角特征融合的网络异常流量检测方法，分别基于时序视角和交互视角对网络流量中数据包的时序关系及交互关系建模，学习其时序特征与交互特征，并将不同视角下的特征融合进行异常流量的检测。在6个真实的流量数据集上进行的大量实验表明，所提方法在网络异常流量检测方面具有优异的性能，弥补了单一视角下检测的不足。

## Network Anomaly Traffic Detection via Multi-view Feature Fusion

SONG Hao[1,2] FU Wentao[1,2] CHEN Xuanze[1,2] JIN Chengxiang[1,2] ZHOU Jiajun[1,2,*] YU Shanqing[1,2] XUAN Qi[1,2]

(1. *Institute of Cyberspace Security, Zhejiang University of Technology, Hangzhou* 310023, *China*;
2. *Binjiang Institute of Artificial Intelligence, Zhejiang University of Technology, Hangzhou* 310056, *China*)

**Abstract** Traditional anomalous traffic detection methods, based on single-view analysis, have obvious limitations in dealing with complex attacks and encrypted communications. To address this, the paper proposes a network anomaly traffic detection method via multi-view feature fusion (MuFF). MuFF models the temporal and interactive relationships of packets in network traffic based on the temporal and interactive viewpoints respectively. It learns temporal and interactive features. These features are then fused from different perspectives for anomaly traffic detection. Extensive experiments on six real traffic datasets show that MuFF has excellent performance in network anomalous traffic detection, which makes up for the shortcomings of detection under a single perspective.

信息化社会快速发展，互联网作为全球信息交流的基础设施，其安全成为维系社会秩序与经济发展的关键。网络流量作为互联网活动的直接反映，不仅承载了海量的用户交互信息，也潜藏着诸多安全隐患。尤其是异常流量的涌现，既可能是黑客入侵的初步信号，也可能是系统内部故障的外在反映，对网络系统的正常运行构成了严重威胁。因此，高效、准确地识别网络流量中的异常现象并及时作出响应，成为维护网络安全的关键。

网络流量异常检测是网络安全领域中一项基础但关键的任务，旨在从网络上传输的数据流量中识别出偏离正常模式的网络行为。这些异常往往与分布式拒绝服务攻击(distributed denial of service，DDoS)、恶意软件传播等网络安全事件密切相关。恶意流量的泛滥能大规模消耗网络资源，引发服务中断，严重时还会导致整个网络基础设施陷入瘫痪状态。恶意流量常搭载木马、勒索软件等恶意负载，利用先进的隐匿技术渗透目标系统，执行高级持续性威胁(advanced persistent threat，APT)，对国家关键信息基础设施构成战略级威胁。因此，攻击方式的不断演化，对网络流量异常的检测准确性、覆盖全面性、响应实时性提出了更高的要求。

面对日益复杂多变的网络环境，网络流量的异常检测技术需应对新的挑战。传统的网络异常流量检测方法，如基于端口号的映射匹配[1]、基于深度包的检测[2]、基于传统机器学习[3-5]和深度学习的检测[6-8]等，虽然在一定程度上能够应对特定类型的异常，但在面对复杂的网络攻击手段、大规模的数据流量以及加密通信等问题时，其检测效率和准确性均受到限制。特别是这些方法往往侧重于从单一视角分析流量特性，存在一定的检测盲区，忽视了网络流量在时间和空间维度上的复杂交互，这无疑降

收稿日期 2024-06-05 录用日期 2024-08-29
国家自然科学基金(62103374, U21B2001)，杭州市重点研发计划(2022C01018, 2024C01025)资助
*通信作者简介 周嘉俊(1995— )，男，浙江杭州人，博士，助理研究员。

低了异常流量的检测效果。

网络异常流量检测旨在通过对流量的监督和管理，及时识别和发现异常的流量行为或模式，是维护网络空间安全的重要手段。由于网络攻击流量具有多样化的特性，攻击行为所产生的流量通常可以被划分为多个类别[9]，由此催生了基于分类的网络异常流量检测方法。通过二分类的方式，网络流量可以被划分为正常或恶意流量；而通过多分类的方式，网络异常流量可以被进一步细分为不同的攻击类型。在此基础上，许多基于统计学习和机器学习的分类模型被应用于网络异常流量检测，但是不同方法从不同视角出发，所捕捉到的模式与特征各不相同。

在早期的互联网环境中，研究者将传输控制协议(transmission control protocol，TCP)或用户数据报协议(user datagram protocol，UDP)的指定端口号与应用协议相互映射匹配[10]，来区分网络流量的类型。此类方法简单高效，但随着端口伪装技术与动态端口分配技术的普及，其检测效果受到了极大的影响。为避免对端口号的过度依赖，同时保证较好的可扩展性，基于深度包的检测技术(deep packet inspection，DPI)随之被提出。该方法通过对数据包的头部信息及负载进行检测，来判断数据包的真实意图和行为模式[2]。随着网络流量的激增，此类方法在对数据包进行逐字节分析的过程中需要消耗大量的时间，增大网络设备运行压力，同时也难以直接分析包含私有加密协议的数据包。文献[11]率先将深度学习技术应用于流量分类，在特征学习和流量异常检测方面展示了显着的性能。文献[12]将数据包长度序列作为输入，采用双向门控循环单元(gated recurrent unit，GRU)进行特征编码，并在AutoEncoder中引入重建机制，以确保学习到的特征的有效性。文献[13]采用在截断数据包字节序列上训练的多任务Transformer，以监督方式分析流量特征。文献[14]提出一种专为网络流量分类而设计的新型预训练状态空间模型，利用单向Mamba架构进行流量序列建模，并取代Transformer以解决效率问题。将网络流量视为计算机设备之间通讯的语言，即由特殊单词构成的流量序列，并利用序列模型来分析网络流量特征。上述方法未能关注到网络中实体间的交互关系。基于图分析的检测方法通常将流量中的数据包视为节点，将数据包间传输关系视为连边，构建流量交互图，并利用图神经网络等图表征学习技术来挖掘潜在的网络拓扑信息和流量交互信息[15-17]。文献[18]通过设计双嵌入层，分别对数据包的包头和负载字节进行编码，并利用PacketCNN获取流量的数据包级表示，构建流量交互图。文献[19-20]引入超图的概念，通过将流量表示为节点，利用KNN算法构建超边，来反映流量的潜在相似性。随着大型语言模型的兴起，NetGPT和Lens等预训练流量基础模型被用来同时解决流量分析和生成问题。

本文从时序视角和交互视角出发来分析网络流量，提出了一种多视角特征融合(multi-view feature fusion，MuFF)的网络异常流量检测方法。在时序视角下，从流量中提取数据包长度、传输方向和负载字节信息，并进行序列建模，形成数据包长度序列和负载字节序列，结合长短期记忆网络(long short-term memory，LSTM)和卷积神经网络(convolutional neural network，CNN)，捕获数据包长度随时间变化的长期依赖性和周期性特征，以及负载中携带的恶意信息如隐匿的恶意指令序列或特殊编码模式。在交互视角下，根据流量中数据包的传输方向和时序分布，构建了分层的数据包交互图，并利用图卷积网络(graph convolutional network, GCN)捕获数据包之间隐含的拓扑关联和相互作用。模型通过融合时序和交互特征，充分发挥不同视角间的互补优势，确保在复杂场景下对异常流量检测的有效性和鲁棒性。

## 1　模型设计

本文从流量中数据包的时序和交互视角出发，提出了一种多视角特征融合的异常流量检测方法。该方法基于原始流量数据分别进行时序建模和交互建模，利用卷积神经网络、时序模型和图神经网络模型从不同视角提取流量特征，用于异常流量检测。该模型设计克服了传统单一视角方法的局限性，通过多视角特征融合，提升对复杂网络行为的理解与异常模式的识别能力，尤其是在对抗日益复杂的网络攻击与隐蔽流量操纵策略中展现出显著优势。整体框架如图1所示。

### 1.1　时序视角感知

许多承载网络攻击的异常流量在时序特征上表现出明显不同于正常流量的模式。例如，网络流量在非高峰时段出现持续上升，偏离了预期模式，可能意味着正在进行的攻击活动或系统故障；或者频繁出现大量相同大小或类型的垃圾数据包，企图占用网络资源。因此，从时序特征感知的视角出发，通过CNN和LSTM深入挖掘流量数据包内潜在的时序行为模式，以增强异常流量检测的时序感知能力。

#### 1.1.1　流量时序建模

基于时序视角进行流量特征感知前，首先从流量数据中提取序列化信息。

对流量中的数据包，提取数据包的传输方向和

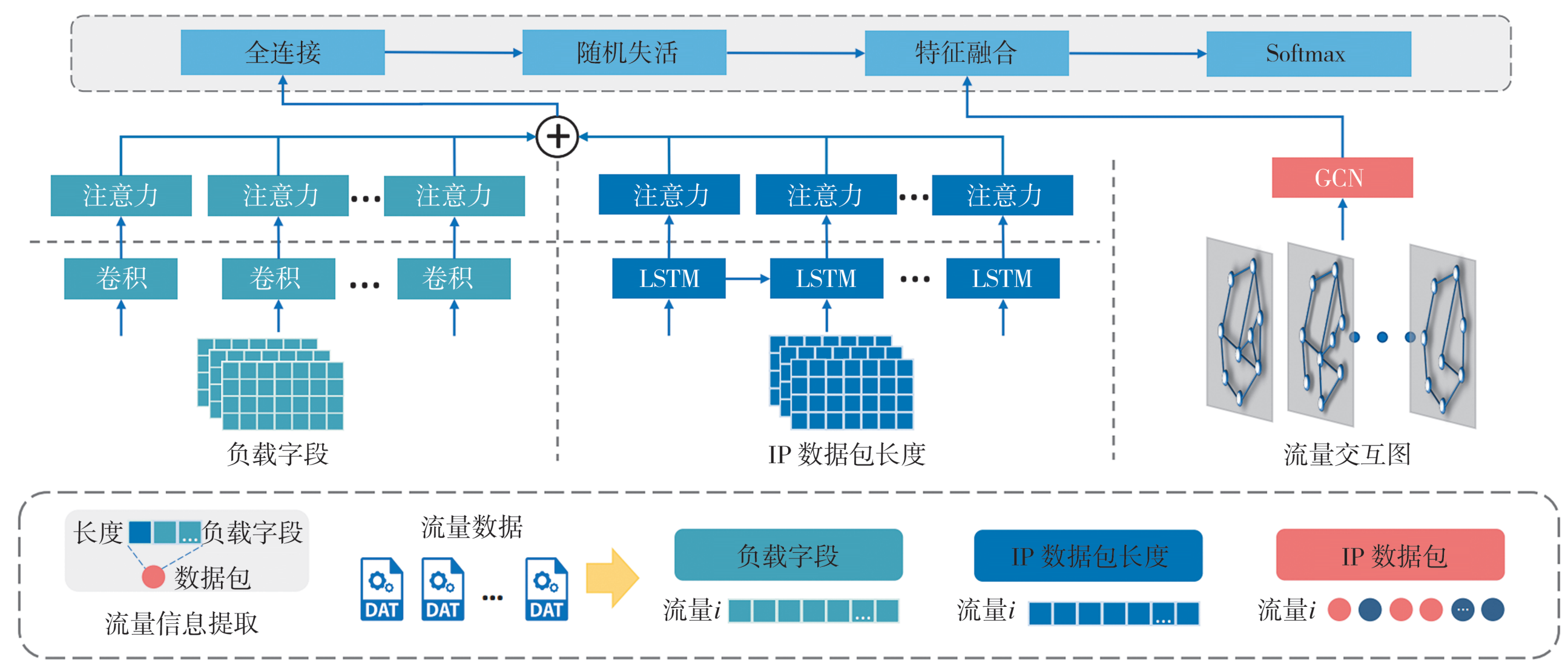

图1　多视角特征融合模型框架

Fig. 1　Multi-view feature fusion model framework

长度信息，并将其按照数据包产生的时间进行排列，构成流量的数据包长度序列。对于长时间持续的网络流量，仅取其前$n$个数据包作为代表，提取其中的信息来构建序列；对于数据包数量不足$n$的网络流量，用0填充缺失部分，最终确保序列数据长度的一致性。例如，考虑一段包含100个数据包的网络流量。选取前$n$=40个数据包作为代表，其长度分别为：100 bytes, 120 bytes, 80 bytes, …, 60 bytes。由于通信过程的双向性，数据包的不同传输方向由±1来表示。最终按照数据包产生的顺序构建长度序列，即[100, −120, 80, …, −60]。对于数据包数量不足40个的情况，则用0填充剩余的位置，以确保序列长度的一致性，即[100, −120, 80, …, 0]。

对数据包中的负载信息，以$m$为阈值，对每个数据包的负载提取前$m$个字节的信息，并将其从16进制转换为10进制，以适应序列数据的格式，最终形成流量的负载字节序列。对于负载内容不足$m$字节的数据包，用0填充缺失部分。例如，从每个数据包中提取前16个字节的信息。其中，第1个数据包的前16个字节为2a0500a800002a0500a900002a0500bc。将这些十六进制数值转换为十进制数值，以适应序列数据的格式，即 [42, 5, 0, 168, 0, 0, 42, 5, 0, 169, 0, 0, 42, 5, 0, 188]。对于有效负载内容不足16 bytes的数据包，则用0填充剩余的位置，以确保序列长度的一致性。

### 1.1.2　流量时序特征提取

针对流量的负载字节序列，利用CNN对其包含的负载内容进行特征提取。CNN通过局部连接和权值共享的特性，能够在负载数据中自动学习并提取出具有特定模式的特征，如特定的恶意指令序列、恶意代码片段或异常的二进制模式。通过组合多层卷积、池化以及非线性激活，CNN能够逐步提炼出数据包负载中的高层次抽象特征，这些特征能够有效反映潜在的恶意活动迹象。

本文所用的CNN模型由两个卷积层构成，在每一个卷积层后都接有归一化层、ReLU激活函数和最大池化层，以此来对抽取的潜在特征进行标准化处理，保留其重要特征并减少内存消耗。卷积层的具体计算公式如下：

$$
\begin{aligned}
H_1 &= \mathrm{MaxPool1d}\left(\mathrm{ReLU}\left(\mathrm{BatchNorm}\left(\mathrm{Conv1d}\left(X_{\mathrm{payload}}\right)\right)\right)\right) \\
H_2 &= \mathrm{MaxPool1d}\left(\mathrm{ReLU}\left(\mathrm{BatchNorm}\left(\mathrm{Conv1d}\left(H_1\right)\right)\right)\right)
\end{aligned}
\tag{1}
$$

其中，Conv1d为一维卷积操作，BatchNorm为归一化操作，MaxPool1d为最大池化操作。在经过两层卷积层后，引入了注意力机制（attention），通过动态地为输入的不同部分分配权重，使得模型能够聚焦于数据中的最重要特征，显著提高模型的识别性能和鲁棒性。利用两个全连接层和一个ReLU激活函数计算注意力权重，并以此对原始特征进行加权操作，该过程表示如下：

$$
\begin{aligned}
e_i &= \sigma\left(\boldsymbol{W}_2 \cdot \mathrm{ReLU}\left(\boldsymbol{W}_1 \cdot h_i\right)\right) \\
a_i &= \frac{e\left(e_i\right)}{\sum_j e\left(e_j\right)} \\
h_{\mathrm{att}} &= \sum_i a_i \cdot h_i
\end{aligned}
\tag{2}
$$

其中，$e$为注意力权重，$\boldsymbol{W}_1$和$\boldsymbol{W}_2$是全连接层的权重矩阵，$\sigma$为sigmoid激活函数，$a$为归一化的注意力系

数。将加权后的特征展平并通过全连接层映射到$M$维特征向量：

$$\begin{aligned} \boldsymbol{h}_{\text{flatten}} &= \text{Flatten}\left(h_{\text{att}}\right) \\ \boldsymbol{Z}_{\text{CNN}} &= \text{Linear}\left(h_{\text{flatten}}\right) \end{aligned} \tag{3}$$

针对流量的数据包长度序列，利用LSTM对其进行特征提取，捕捉长度序列的长期依赖性和周期性变化。LSTM通过其独特的门控机制，包括输入门、输出门和遗忘门，能够有效地记住过去的流量模式，遗忘无关信息，预测未来流量的正常变化趋势，从而准确地区分出偏离正常模式的异常流量行为。模型更新公式如下：

$$\begin{aligned} i_t &= \sigma\left(\boldsymbol{W}_i \cdot \left[H_{t-1}, x_t\right] + b_i\right) \\ o_t &= \sigma\left(\boldsymbol{W}_o \cdot \left[H_{t-1}, x_t\right] + b_o\right) \\ f_t &= \sigma\left(\boldsymbol{W}_f \cdot \left[H_{t-1}, x_t\right] + b_f\right) \\ C_t &= f_t \otimes C_{t-1} + i_t \otimes \tanh\left(\boldsymbol{W}_c \cdot \left[H_{t-1}, x_t\right] + b_c\right) \\ H_t &= o_t \otimes \tanh\left(C_t\right) \end{aligned} \tag{4}$$

其中，$\sigma$为sigmoid激活函数，tanh为双曲正切函数，$\boldsymbol{W}$为对应的权重，$b$为偏置。在LSTM层，同样引入了注意力机制以增强其学习效果。LSTM得到的输出$H_t$输入到注意力层，得到输出向量：

$$\boldsymbol{Z}_{\text{LSTM}} = \text{Attention}\left(H_t\right) \tag{5}$$

## 1.2 交互视角感知

部分异常流量在数据包的交互上表现出明显区别于正常流量的行为模式。例如，某一IP地址在短时间内对多个不同端口发起大量连接请求；或者某一段时间内频繁出现特定协议的数据包，这些行为可能暗示正在进行的扫描攻击或恶意活动。因此，从数据包的交互视角出发，通过深入挖掘流量数据包内潜在的交互行为模式，以增强异常流量检测模型的交互关系感知能力。

### 1.2.1 流量交互建模

基于交互视角进行流量特征感知前，首先从流量数据中提取交互信息。采用一种流量构图策略，通过构建流量数据包交互图来表示流量内部数据包之间的交互关系[21]。服务器与客户端之间的流量交互如图2所示，将数据从客户端发送至服务器的方向标记为负，将服务器发往客户端的方向标记为正。这一定义考虑到通信过程的双向性，有助于更准确地理解和分析网络流量的动态特征。图3展示了一个流量交互图的示例，将数据包作为节点，将其长度和方向作为节点特征，将数据包之间的通信关系作为连边，不同方向的数据包用不同颜色标记。同时，引入了层的概念，将连续的同方向数据包囊括在一层中（图中以列表示）。

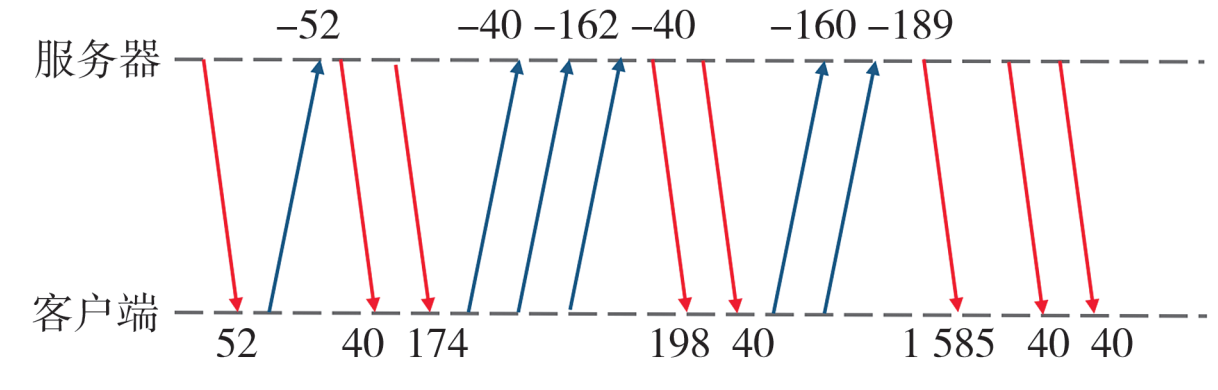

图2 服务器与客户端之间的流量交互

Fig. 2 Traffic interaction between the server and the client

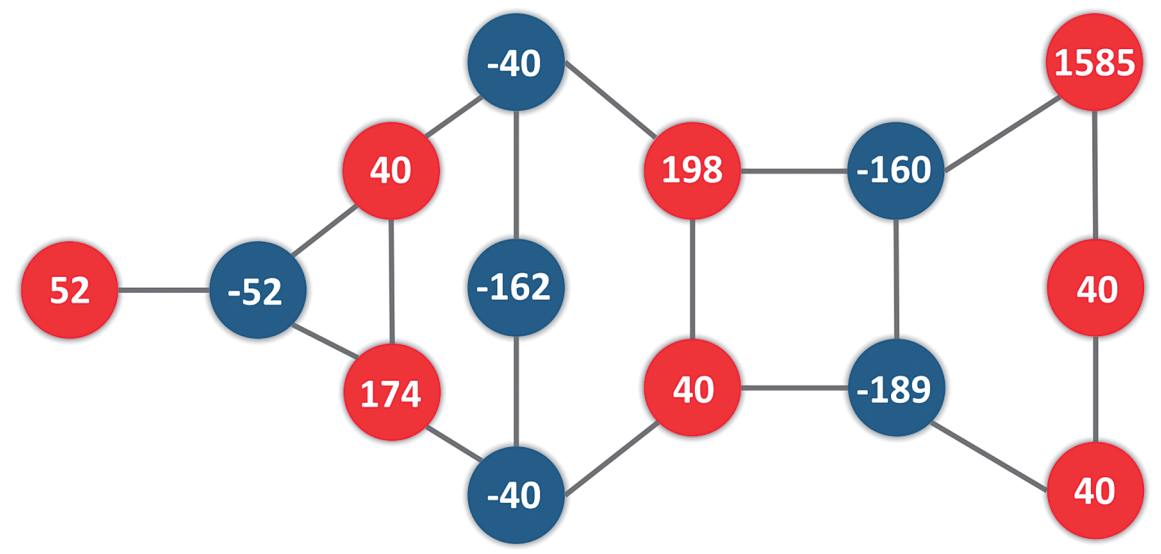

图3 流量数据包交互图

Fig. 3 Traffic packet interaction graph

考虑到不同网络流量的持续时间差异，针对持续时间较长的网络流量采取了一种特殊处理方式。选择长时流量的前$n$个数据包来代表整个流量，并以这$n$个数据包为基础构建流量数据包交互图。这一策略的优势在于，通过提前处理，能够更早地检测到持续时间较长的恶意流量，并且及时采取防御措施。这种方法不仅节省了计算资源，同时也提高了对网络流量的分析效率和准确性。

### 1.2.2 流量交互特征提取

针对流量数据包交互图，采用图卷积网络GCN模型[22]来捕捉网络流量中数据包的复杂传输关系和潜在的异常交互模式。GCN模型通过直接在图结构上进行特征的传播、聚合和转换操作，能够有效学习节点特征及其邻居节点间的相互作用，不仅能够揭示隐藏的攻击路径和协同行为，还能捕捉到异常的流量分布和通信模式。

本文中的GCN模型用于学习每一个流量数据包交互图的表征信息。在该模型中，将流量数据包交互图的邻接信息和节点特征作为输入，得到整个交互图的表示作为输出。在对邻域节点信息进行聚合之前，先将流量交互图的邻接信息进行归一化处理：

$$\tilde{\boldsymbol{A}}_{\text{norm}} = \tilde{\boldsymbol{D}}^{-1/2} \tilde{\boldsymbol{A}} \tilde{\boldsymbol{D}}^{-1/2} \tag{6}$$

其中，$\tilde{\boldsymbol{A}} = \boldsymbol{A} + \boldsymbol{I}_n$表示对邻接矩阵添加自环，$\tilde{\boldsymbol{D}}$是$\tilde{\boldsymbol{A}}$的度矩阵，有$\tilde{\boldsymbol{D}}_{ii} = \sum_j \tilde{\boldsymbol{A}}_{ij}$。GCN对节点特征的传播和更新的过程可以表示为

$$\boldsymbol{H}^{(l+1)} = \sigma\left(\tilde{\boldsymbol{A}}_{\text{norm}} \boldsymbol{H}^{(l)} \boldsymbol{W}^{(l)}\right) \tag{7}$$

式（7）表示第$l$层的节点特征$\boldsymbol{H}^{(l)}$通过邻居节点

的信息聚合和权重矩阵 $\boldsymbol{W}^{(l)}$ 的线性变换，再经过非线性激活函数 $\sigma$，得到第 $l+1$ 层的节点特征 $\boldsymbol{H}^{(l+1)}$。在经过 $k$ 次特征传播和更新后，可以得到流量数据包交互图中的节点信息 $\boldsymbol{H}^{(k)}$。为了获得整个交互图的全局信息，对所有节点的特征进行全局平均池化：

$$\boldsymbol{Z}_{\mathrm{GCN}} = \mathrm{global_mean_pool}\left(\boldsymbol{H}^{(k)}\right) \tag{8}$$

### 1.3　多视角特征融合与检测

在多视角特征融合阶段，首先将序列视角下感知到的数据包长度序列特征和负载字节序列特征进行拼接，并将拼接后的特征输入到全连接层中，以更好地拟合数据分布。同时，借助Dropout以减少模型过拟合风险：

$$\boldsymbol{Z}_{\mathrm{SEQ}} = \mathrm{Linear}\left(\mathrm{Dropout}\left(\mathrm{Linear}\left(\boldsymbol{Z}_{\mathrm{CNN}} \mathbin{/\!/} \boldsymbol{Z}_{\mathrm{LSTM}}\right)\right)\right) \tag{9}$$

其中，$/\!/$ 表示拼接操作。在对序列视角下感知的特征进行融合后，进一步将交互视角下感知到的数据包交互特征与序列特征进行融合，得到网络流量最终的特征表示：

$$\boldsymbol{Z} = \alpha \cdot \boldsymbol{Z}_{\mathrm{GCN}} + (1-\alpha)\boldsymbol{Z}_{\mathrm{SEQ}} \tag{10}$$

使用全连接层来进行线性变换，并使用Softmax函数对目标流量进行分类：

$$\boldsymbol{Y}_i = \mathrm{Softmax}(\boldsymbol{Z}) \tag{11}$$

本文使用了交叉熵损失函数作为训练损失函数，模型通过最小化损失函数，实现对其参数的优化，损失函数公式如下：

$$L = -\frac{1}{N}\sum_{i=1}^{N}\sum_{k=1}^{K} y_{i,k}\log\left(\hat{y}_{i,k}\right) \tag{12}$$

其中，$N$ 表示总样本的数量，$K$ 表示样本中的类别数，$y_{i,k}$ 是第 $i$ 个样本在第 $k$ 类标签的真实值，$\hat{y}_{i,k}$ 是模型预测的第 $i$ 个样本属于第 $k$ 类的概率。

## 2　实验验证

为了评估本文所提出模型的性能，在6个广泛使用的公开数据集上进行网络流量异常检测实验。

### 2.1　数据集概述

为验证所提方法在网络异常流量检测上的有效性和优越性，在6个公开的真实网络流量数据集上进行了实验。

CIC-IOMT2024：该数据集由加拿大网络安全研究所发布，用于维护医疗物联网（internet of medical things，IoMT）领域安全。它基于一个由40台IoMT设备（25台真实设备和15台模拟设备）组成的测试平台，实施了5类共18种攻击，全方位覆盖了医疗领域中常见的通信协议，包括但不限于Wi-Fi、MQTT协议及蓝牙技术。数据收集过程中运用网络嗅探技术，在交换机与各IoMT设备之间捕获流量数据，为研究提供了丰富且真实的数据资源。

UNSW-NB15[23-26]：新南威尔士大学堪培拉分校的Cyber Range实验室于2015年使用IXIA PerfectStorm工具生成的原始网络数据包，包含真实的网络正常活动和攻击行为。该数据集包含9种攻击类型。UNSW-NB15数据集的异常行为更加新颖和均衡，适合用于网络异常流量检测研究。

Darknet2020[27]：该数据集由两个公开的加密数据集（ISCX-VPN2016和ISCX-Tor2016）合并而成一个完整的暗网流量数据集。其中，加密流量分别涵盖Tor和VPN两类不同加密通信方式，包含6种不同的加密应用程序流量。暗网流量分类对于实时应用程序的分类非常重要。分析暗网流量有助于在恶意软件肆虐前对其进行早期监控，并在爆发后对恶意活动进行检测。

ISCX-VPN2016[28]：该数据集包含通过虚拟专用网络（virtual private network，VPN）隧道传输的加密通信流量，涵盖网络电话、电子邮件等5种不同的应用程序。VPN通常用于规避审查或通过协议混淆隐藏位置，常用于访问被封锁的网站或服务，因此，VPN流量的识别与检测也是一项极具挑战性的工作。

CIC-IoT2023[29]：该数据集提供了物联网环境中大规模攻击的实时数据。加拿大网络安全研究中心在一个由105台真实物联网设备组成的复杂拓扑中执行了7大类攻击：分布式拒绝服务、DoS、侦察、基于Web的攻击、暴力破解、欺骗以及Mirai攻击。这些攻击均由恶意物联网设备发起，并针对网络中的其他物联网设备实施。

USTC-TFC2016[30]：该数据集由中国科学技术大学团队发布，包含由恶意软件和良性应用程序组成的加密流量。其中，恶意流量与Cridex、Geodo、Htbot等10种恶意病毒或木马相关。

### 2.2　实验对比方法

将MuFF与以下方法进行比较以评估其流量异常检测的性能。

GraphSAGE[31]：通过引入分层采样策略来限制每层聚合邻居信息的数量，并支持多种聚合操作如平均、最大池化等，灵活地汇总邻居节点的特征，生成节点的嵌入向量，这些向量充分编码了节点的本地结构特征及全局上下文信息。

GIN[32]：图同质化网络通过灵活的聚合函数，将流量交互图中的邻居节点信息与中心节点特征相结合，通过可学习的参数权重来线性或非线性地累加邻居特征，从而在保持图同构不变性的同时，增强

模型的表达能力。

GAT[33]：图注意力网络将注意力机制融入到图数据的学习过程中，从而动态地赋予图中相邻节点不同的权重。每个节点的特征表示不仅依赖于其邻居节点的特征，还取决于一个自适应学习的注意力系数，这个系数能够反映当前节点与其各邻居节点间关系的重要性。

MuFF-CNN：该方法为MuFF单时序视角的变体方法，只利用CNN来学习网络流量的负载序列中所具有的特定模式。

MuFF-LSTM：该方法为MuFF单时序视角的另一种变体方法，只利用LSTM捕捉数据包长度随时间变化的长期依赖性和周期性特征。

MuFF-GCN：该方法为MuFF单交互视角的变体方法，只利用GCN在流量交互图的节点特征间传播和变换信息，最终得到整个图的嵌入表示来学习流量交互图的特征表示。

E-GraphSAGE[34]：将网络中的通信端点(由IP地址和端口号表示)映射为图节点，网络流量映射为边。该方法基于GraphSAGE进行扩展，其消息传递函数不仅考虑节点特征，还考虑边特征。通过采样和聚合图中的边信息来生成边嵌入，进而实现对恶意网络流量的有效检测。

GraphDDoS[35]：该方法通过将网络流量转换为端点流量图来捕捉数据包之间的关系(单流结构)和多个流之间的关系，借助基于GNN的消息传递网络进行节点表示学习，并通过聚合节点特征获得整个图的表示，以区分网络中的DDoS攻击流量和正常流量。

## 2.3 实验超参数设置

本文提出的算法基于PyTorch及其对应的PyTorch Geometric库实现。在交互特征提取中，使用了3层GCN模型，并将隐层维度设置为512维，对应于图嵌入的维度。同时，为了优化神经网络，防止过拟合问题的出现，在3层图卷积层之间加入了Dropout机制，并将其参数设置为0.5。在时序特征提取中，结合了LSTM模型与CNN模型，在LSTM层中，采用了与GCN相同的设置。除此之外，使用了ReLU激活函数作为模型的非线性变换函数，并使用Softmax函数作为分类器。在优化器与损失函数选择方面，选择了Adam优化器，并将其学习率设置为0.002以执行反向传播阶段的梯度下降，同时选择交叉熵损失作为损失函数。其他超参数设置如表1所示。

表1 超参数设置

Table 1 Hyperparameter settings

| 超参数 | 描述 | 设置 |
|---|---|---|
| $n$ | 一个流的最大数据包数量 | 40 |
| $m$ | 每个数据包提取的负载字节数 | 16 |
| $\alpha$ | 图模型与序列模型融合的权重参数 | 0.5 |
| Kernelsize1 | 第1层卷积层的卷积核大小 | 25 |
| Kernelsize2 | 第2层卷积层的卷积核大小 | 25 |
| Kernelsize3 | 池化层窗口大小 | 3 |
| Strid1 | 第1层卷积层的步长 | 1 |
| Strid2 | 第2层卷积层的步长 | 1 |
| Strid3 | 池化层窗口步长 | 3 |
| Padding1 | 第1层卷积层填充大小 | 12 |
| Padding2 | 第2层卷积层填充大小 | 12 |
| Padding3 | 池化层填充大小 | 1 |

## 2.4 模型评价指标

本实验采用准确率(accuracy)、宏平均精确率(macro precision)、宏平均召回率(macro recall)、宏平均$F_1$分数(Macro $F_1$-Score)作为评价指标，各指标的计算方法如下：

准确率描述了正确预测的样本数与总样本数的比值。

$$A_{\mathrm{C}} = \frac{S_{\mathrm{TP}} + S_{\mathrm{TN}}}{S_{\mathrm{TP}} + S_{\mathrm{TN}} + S_{\mathrm{FP}} + S_{\mathrm{FN}}} \tag{13}$$

宏平均精确率是针对每个类别分别计算精确率$Pr_{\mathrm{i}}$，然后取所有类别的精确率的平均值，同样不考虑不同类别中正例的数量，强调每个类别单独的表现。

$$\begin{gathered} Pr_{\mathrm{i}} = \frac{S_{\mathrm{TP}_i}}{S_{\mathrm{TP}_i} + S_{\mathrm{FP}_i}} \\ M-Pr = \frac{1}{C}\sum_{i=1}^{C} Pr_{\mathrm{i}} \end{gathered} \tag{14}$$

类似地，宏平均召回率是针对每个类别单独计算召回率$Re_{\mathrm{i}}$，再对所有类别的召回率求平均。这确保了每个类别异常情况的检测都得到了平等的重视，不管其在数据集中占比如何。

$$\begin{gathered} Re_{\mathrm{i}} = \frac{S_{\mathrm{TP}_i}}{S_{\mathrm{TP}_i} + S_{\mathrm{FN}_i}} \\ M-Re = \frac{1}{C}\sum_{i=1}^{C} Re_{\mathrm{i}} \end{gathered} \tag{15}$$

宏平均$F_1$分数是每个类别的$F_1$分数的平均值。$F_1$分数是精确率和召回率的调和平均，它在评估类别不平衡数据时特别有效，因为宏平均$F_1$同时考虑了精确率和召回率，且对每个类别给予同等权重。

$$F_{1i} = 2*\frac{Pr_{\mathrm{i}}*Re_{\mathrm{i}}}{Pr_{\mathrm{i}} + Re_{\mathrm{i}}}$$

$$M-F_1-\text{Score}=\frac{1}{C}\sum_{i=1}^{C}F_{1i} \tag{16}$$

这些公式中，$S_{\text{TP}}$(true positive)表示真正例，反映了模型在所有实际为正类样本中，成功识别出正类的能力；$S_{\text{TN}}$(true negative)表示真负例，表明模型在所有实际为负类的样本中，正确识别出负类的能力；$S_{\text{FP}}$(false positive)表示假正例，衡量了模型将负类错误地标记为正类的频率，过高可能会导致误报增多，降低模型的精确度；$S_{\text{FN}}$(false negative)表示假负例，衡量了模型未能识别出正类的错误次数。

## 3　实验结果

本文将所提出的方法与网络异常流量检测中常用的方法GraphSAGE、GIN、GAT、E-GraphSAGE、GraphDDoS进行了对比，分别考察其在多个数据集上的准确率(accuracy)、宏平均精确率(macro Precision)、宏平均召回率(macro Recall)、宏平均$F_1$分数(macro $F_1$-Score)4种指标的表现，结果如表2和表3所示。实验结果显示，提出的MuFF模型在上述6个数据集上的性能均显著优于其他对比模型。在Darknet2020和CIC-IoT2023数据集上，MuFF在所有指标上均表现出色，特别是在宏平均精度和召回率上，与次优结果相比，最高提升率分别达到1.69%和2.86%。在USTC-TFC2016数据集上，MuFF的表现更加优异，准确率和宏平均召回率分别达到了97.67%和97.15%。尽管MuFF在CIC-IOMT2024、UNSW-NB15和ISCX-VPN2016数据集上的部分指标值略低于对比方法，但其在其他指标上的优势仍显示了其在异常流量检测任务中的有效性和鲁棒性。

表4~表9进一步展示了MuFF在6个数据集上对不同类型流量的检测效果。在CIC-IOMT2024数据

表2　CIC-IOMT2024、UNSW-NB15和Darknet2020数据集上不同模型性能比较

Table 2　Comparison of different model performances on the CIC-IOMT2024, UNSW-NB15, and Darknet2020 datasets

| 模型 | 准确率 | | | 宏平均精确率 | | | 宏平均召回率 | | | 宏平均$F_1$分数 | | |
|---|---|---|---|---|---|---|---|---|---|---|---|---|
| | CIC-IOMT2024 | UNSW-NB15 | Darknet2020 | CIC-IOMT2024 | UNSW-NB15 | Darknet2020 | CIC-IOMT2024 | UNSW-NB15 | Darknet2020 | CIC-IOMT2024 | UNSW-NB15 | Darknet2020 |
| GraphSAGE | 86.51±3.81 | 91.95±0.07 | 72.16±0.21 | 54.87±2.28 | 57.94±1.01 | 25.34±267 | 52.88±1.86 | 51.58±0.59 | 19.74±1.56 | 53.02±2.04 | 51.54±0.41 | 18.17±0.97 |
| GIN | 73.59±9.45 | 91.80±0.16 | 90.60±0.31 | 44.72±3.15 | 67.22±9.99 | 72.56±2.54 | 40.24±5.59 | 51.02±0.92 | 56.74±0.42 | 38.64±7.19 | 52.21±1.17 | 59.77±1.51 |
| GAT | 82.78±9.41 | 91.59±0.54 | 72.24±0.23 | 57.14±2.69 | 59.04±0.94 | 25.34±1.59 | 50.85±4.87 | 50.32±1.01 | 20.26±4.27 | 51.11±5.61 | 51.09±1.03 | 19.04±3.03 |
| MuFF-GCN | 90.63±0.29 | 91.84±0.04 | 89.75±0.32 | 82.15±1.50 | 55.65±1.40 | 76.41±4.44 | 68.13±1.55 | 50.43±0.23 | 53.16±2.03 | 72.22±1.47 | 50.33±0.56 | 57.10±1.04 |
| MuFF-CNN | 93.78±0.27 | 92.25±0.35 | 93.85±0.14 | 88.49±2.00 | 63.46±1.30 | 81.81±0.24 | 84.45±0.43 | 62.30±0.44 | 77.28±1.34 | 86.03±0.92 | 62.46±0.64 | 79.06±1.00 |
| MuFF-LSTM | 94.05±0.13 | 92.98±0.12 | 94.04±0.02 | 88.72±0.44 | 66.42±0.28 | 80.42±1.55 | 84.66±0.60 | 65.41±1.30 | 78.25±0.09 | **86.43±0.56** | 64.76±0.86 | 79.11±0.91 |
| CNN+GCN | 93.57±0.24 | 92.52±0.24 | 93.61±0.06 | 87.30±2.08 | 64.29±2.75 | 84.43±1.11 | 83.89±1.09 | 61.90±2.49 | 78.68±0.47 | 84.86±1.16 | 60.43±1.06 | 78.66±0.91 |
| LSTM+GCN | 93.28±0.36 | 91.94±0.32 | 92.82±0.43 | 89.36±3.82 | **75.35±3.99** | 80.70±2.20 | 81.80±3.26 | 56.55±1.56 | 75.00±0.82 | 83.42±1.96 | 54.54±3.34 | 75.40±1.44 |
| CNN+LSTM | 93.79±0.40 | 92.18±0.55 | 94.11±0.39 | 88.37±2.61 | 63.89±0.87 | 82.22±0.77 | 83.69±1.27 | 62.74±3.34 | 78.47±1.10 | 85.61±1.75 | 62.65±1.84 | 79.75±0.96 |
| E-GraphSAGE | 85.83±4.71 | 73.62±1.95 | 66.27±2.24 | 42.90±2.77 | 43.02±9.14 | 35.58±7.34 | 46.44±2.43 | 27.62±2.83 | 25.33±2.52 | 44.51±2.95 | 28.73±4.58 | 24.93±3.41 |
| GraphDDoS | 90.35±0.06 | 91.94±0.23 | 91.22±0.23 | 70.45±9.61 | 57.50±1.26 | 71.29±0.52 | 60.75±1.29 | 54.09±3.61 | 67.32±0.88 | 62.14±2.32 | 54.56±2.52 | 68.66±0.80 |
| MuFF | **94.21±0.15** | **93.11±0.28** | **94.83±**0.23 | **89.38±1.09** | 75.21±4.12 | **85.80±0.89** | **85.34±1.26** | **69.48±2.90** | **81.33±0.79** | 86.19±1.20 | **67.87±1.44** | **81.87±0.47** |

表3 ISCX-VPN2016、CIC-IoT2023和USTC-TFC2016数据集上不同模型性能比较

Table 3 Comparison of different model performances on the ISCX-VPN2016, CIC-IoT2023, and USTC-TFC2016 datasets

| 模型 | 准确率 | | | 宏平均精确率 | | | 宏平均召回率 | | | 宏平均$F_1$分数 | | |
|---|---|---|---|---|---|---|---|---|---|---|---|---|
| | ISCX-VPN2016 | CIC-IoT2023 | USTC-TFC2016 | ISCX-VPN2016 | CIC-IoT2023 | USTC-TFC2016 | ISCX-VPN2016 | CIC-IoT2023 | USTC-TFC2016 | ISCX-VPN2016 | CIC-IoT2023 | USTC-TFC2016 |
| GraphSAGE | 67.04±0.26 | 50.24±0.23 | 85.09±0.71 | 42.87±3.25 | 24.90±1.59 | 43.45±3.09 | 31.26±0.96 | 24.65±0.17 | 41.11±0.99 | 33.00±1.16 | 23.19±0.22 | 39.88±1.01 |
| GIN | 75.29±1.22 | 58.80±0.69 | 97.67±0.18 | 65.73±6.46 | 52.68±5.58 | 91.61±0.59 | 54.80±4.33 | 35.99±1.16 | 89.21±0.21 | 57.09±4.51 | 37.94±1.29 | 90.10±0.34 |
| GAT | 66.17±0.21 | 49.16±0.29 | 84.07±0.54 | 33.29±0.92 | 15.35±2.51 | 31.43±1.34 | 28.21±2.29 | 20.33±1.65 | 29.99±1.32 | 28.39±1.94 | 16.57±1.76 | 27.98±0.27 |
| MuFF-GCN | 75.64±0.25 | 58.37±1.47 | 95.78±0.19 | 65.20±2.19 | 53.42±3.42 | 88.97±0.33 | 51.49±3.20 | 34.35±1.91 | 85.75±0.66 | 55.83±2.55 | 35.69±2.85 | 86.99±0.56 |
| MuFF-CNN | 84.06±0.09 | 78.34±0.29 | 97.60±0.08 | 80.95±0.52 | 69.04±1.44 | **96.72±0.10** | 74.63±0.41 | 59.02±0.34 | 96.82±0.28 | 76.67±0.79 | 62.42±0.43 | 96.76±0.14 |
| MuFF-LSTM | 83.06±0.59 | 79.11±0.17 | 97.17±0.21 | 79.11±2.26 | 71.50±0.44 | 95.66±0.16 | **78.40±2.74** | 58.61±0.19 | 95.95±0.47 | 78.04±0.02 | 62.90±0.11 | 95.78±0.19 |
| CNN+GCN | 83.44±0.88 | 77.96±0.27 | 97.43±0.21 | 80.17±1.68 | 80.17±1.68 | 96.68±0.27 | 58.11±0.36 | 58.11±0.36 | 96.67±0.25 | 61.22±0.74 | 61.22±0.74 | 96.60±0.42 |
| LSTM+GCN | 78.50±3.16 | 78.82±0.25 | 96.83±0.43 | 78.20±3.17 | 75.84±0.79 | 94.83±0.85 | 66.58±4.21 | 56.62±0.30 | 95.37±0.61 | 68.99±2.92 | 60.57±0.42 | 94.97±0.88 |
| CNN+LSTM | 83.77±0.07 | 78.23±0.42 | 97.36±0.20 | 78.72±1.76 | 66.58±0.61 | 96.42±0.25 | 75.51±1.53 | 58.80±0.77 | 96.41±0.36 | 75.94±0.82 | 61.57±0.76 | 96.40±0.29 |
| E-Graph SAGE | 81.58±9.21 | 45.32±1.49 | 52.28±6.14 | 54.84±4.68 | 25.09±4.22 | 28.69±4.37 | 41.79±1.13 | 17.23±1.15 | 29.46±5.51 | 44.36±3.36 | 15.90±2.03 | 25.59±4.86 |
| GraphDDoS | 75.27±1.28 | 75.62±0.79 | 93.28±0.37 | 71.90±0.90 | 71.37±1.19 | 88.85±0.54 | 58.92±3.19 | 51.59±0.28 | 87.50±1.46 | 59.36±1.21 | 55.89±0.49 | 87.13±0.58 |
| MuFF | **85.06±0.22** | **79.51±0.36** | **97.67±0.14** | **85.92±1.58** | **79.08±0.76** | 96.55±0.59 | 78.24±2.50 | **60.68±0.62** | **97.15±0.35** | **78.64±2.35** | **63.61±0.14** | **96.87±0.33** |

表4 CIC-IOMT2024数据集上流量多分类结果

Table 4 Traffic multi-classification results on the CIC-IOMT2024 dataset

| 流量类别 | 宏平均精确率 | 宏平均召回率 | 宏平均$F_1$分数 |
|---|---|---|---|
| 良性流量 | 95.42±0.74 | 98.66±0.32 | 96.06±0.32 |
| ARP攻击 | 83.91±4.88 | 89.57±4.17 | 84.11±1.99 |
| 分布式拒绝服务 | 92.79±0.13 | 98.17±0.27 | 95.02±0.09 |
| 拒绝服务 | 99.88±0.09 | 93.77±3.82 | 95.19±0.05 |
| MQTT协议流量 | 97.95±1.04 | 84.04±0.33 | 87.83±1.85 |
| 探查流量 | 79.82±6.67 | 65.52±8.70 | 60.79±1.33 |

表5 UNSW-NB15数据集上流量多分类结果

Table 5 Traffic multi-classification results on the UNSW-NB15 dataset

| 流量类别 | 宏平均精确率 | 宏平均召回率 | 宏平均$F_1$分数 |
|---|---|---|---|
| 良性流量 | 99.68±0.16 | 99.76±0.04 | 99.62±0.06 |
| 拒绝服务 | 40.74±42.87 | 13.33±4.73 | 16.91±4.90 |
| 漏洞利用 | 85.27±1.64 | 97.67±1.08 | 88.03±0.74 |
| 杂项异常 | 72.80±4.85 | 78.71±6.22 | 62.88±3.85 |
| Shellcode 注入 | 76.04±6.96 | 76.65±4.05 | 65.39±0.91 |

集上，MuFF在检测正常流量和DDoS攻击方面表现卓越，这反映了MuFF在处理大比例数据时的鲁棒性

表6　Darknet2020数据集上流量多分类结果

Table 6　Traffic multi-classification results on the Darknet2020 dataset

| 流量类别 | 宏平均精确率 | 宏平均召回率 | 宏平均$F_1$分数 |
|---|---|---|---|
| 网页浏览 | 98.32±0.56 | 98.84±0.64 | 96.01±0.33 |
| 即时通讯 | 76.94±6.82 | 60.71±6.28 | 44.28±2.75 |
| 电子邮件 | 95.03±1.63 | 85.38±1.39 | 84.93±0.92 |
| 文件传输 | 97.82±0.47 | 95.04±0.70 | 94.52±0.27 |
| 点对点流量 | 99.53±0.07 | 99.41±0.11 | 99.27±0.04 |
| 流媒体 | 87.94±3.75 | 86.67±1.95 | 76.00±1.59 |
| 网络电话 | 89.66±3.09 | 87.28±3.22 | 79.95±0.68 |

表7　ISCX-VPN2016数据集上流量多分类结果

Table 7　Traffic multi-classification results on the ISCX-VPN dataset

| 流量类别 | 宏平均精确率 | 宏平均召回率 | 宏平均$F_1$分数 |
|---|---|---|---|
| 即时通讯 | 82.42±4.90 | 95.97±1.06 | 77.10±0.30 |
| 电子邮件 | 98.25±2.48 | 45.74±9.17 | 59.05±5.03 |
| 文件传输 | 88.40±2.08 | 76.92±5.44 | 76.74±2.47 |
| 点对点流量 | 95.42±1.94 | 87.37±5.16 | 86.47±3.30 |
| 流媒体 | 89.93±2.29 | 92.28±0.44 | 86.27±1.47 |
| 网络电话 | 96.82±0.15 | 99.30±0.36 | 90.09±0.04 |

表8　CIC-IoT2023数据集上流量多分类结果

Table 8　Traffic multi-classification results on the CIC-IoT2023 dataset

| 流量类别 | 宏平均精确率 | 宏平均召回率 | 宏平均$F_1$分数 |
|---|---|---|---|
| 分布式拒绝服务 | 79.33±058 | 95.38±0.81 | 84.77±0.06 |
| 漏洞扫描 | 33.33±47.14 | 15.60±2.25 | 18.68±1.92 |
| 良性流量 | 98.73±0.18 | 92.57±0.62 | 94.80±0.24 |
| DNS欺骗 | 92.38±3.24 | 61.37±3.36 | 67.52±1.07 |
| 中间人攻击 | 84.83±1.62 | 73.26±2.93 | 69.54±0.23 |
| 浏览器劫持 | 88.57±6.28 | 38.68±2.42 | 46.06±1.78 |
| 暴力破解 | 75.79±2.33 | 77.34±1.87 | 70.77±0.38 |
| 注入攻击 | 79.35±0.92 | 53.18±0.27 | 61.35±0.42 |

表9　USTC-TFC2016数据集上流量多分类结果

Table 9　Traffic multi-classification results on the USTC-TFC2016 dataset

| 流量类别 | 宏平均精确率 | 宏平均召回率 | 宏平均$F_1$分数 |
|---|---|---|---|
| 良性流量 | 99.96±0.03 | 99.97±0.02 | 99.99±0.01 |
| Cridex木马 | 99.99±0.01 | 100.00±0.00 | 99.99±0.01 |
| Geodo木马 | 99.74±0.30 | 99.87±0.05 | 99.77±0.08 |
| Htbot僵尸网络 | 99.57±0.45 | 99.61±0.35 | 99.23±0.06 |
| Miuref后门 | 99.76±0.19 | 99.97±0.05 | 99.88±0.06 |
| Neris木马 | 97.72±1.61 | 94.29±1.85 | 91.31±0.73 |
| Nsis-ay恶意软件 | 98.92±0.18 | 96.51±0.79 | 94.51±0.67 |
| Shifu木马 | 98.49±0.64 | 99.34±0.47 | 97.96±0.67 |
| Tinba木马 | 97.67±3.29 | 97.47±2.58 | 99.50±0.70 |
| Virut病毒 | 89.80±2.49 | 95.29±0.64 | 87.23±0.68 |
| Zeus木马 | 99.61±0.28 | 98.90±0.45 | 98.17±0.26 |

和高效性。对于较少见的ARP和MQTT攻击，MuFF也能较好地检测。对于Recon攻击，MuFF的检测效果不如其他类型，这表明Recon攻击的特征可能较为隐蔽，模型难以准确识别。在UNSW-NB15、CIC-IoT2023和USTC-TFC2016数据集上，MuFF对正常流量(Benign)的检测表现尤为突出，显示了模型在处理大部分数据时的高效性。对于占比较小的流量如DOS攻击，Vulnerability，MuFF的检测效果较差，表明模型在处理少量数据时能力有限。相比之下，对于Exploits攻击，尽管其数据量也不大，但是MuFF仍表现出色，这可能归因于Exploits攻击特征的明显性。在Darknet2020、ISCX-VPN2016数据集中，MuFF对于不同应用产生的加密流量依然有较好的分类效果，凸显了模型具有良好的泛化性能。综合来看，MuFF在6个数据集上的实验结果展示了其在正常流量和高比例攻击流量(如DDoS和Exploits)检测中的优异表现，但在处理数据量较小或特征不明显的攻击(如DOS和Recon)时，仍有改进空间。这表明未来可以通过增强模型处理小样本的能力来增强其检测的全面性。

## 4　消融分析

为深入分析各视角对模型整体性能的影响，本

文进行了消融分析。从表2可见：1）仅从交互视角出发进行异常流量检测（对应GCN的结果），效果低于时序视角的检测结果。这主要是由于图神经网络模型在实现优异的表征学习效果时，依赖于良好的初始特征构建。然而在构建流量数据包交互图时，仅使用了数据包长度和传输方向作为初始特征，因此，效果不理想是可以理解的；2）仅从时序视角出发进行异常流量检测（对应MuFF-CNN、MuFF-LSTM、CNN+LSTM的结果），取得了较好的效果。这是因为时序模型（如LSTM）能够有效捕捉数据包长度随时间变化的长期依赖性和周期性特征，而卷积神经网络则擅长从负载字节序列中提取高层次特征。这种组合使得模型在检测包含特定模式或特征的恶意流量时表现尤为突出；3）将时序视角和交互视角相结合（对应CNN+GCN、LSTM+GCN、MuFF的结果），效果比单视角更加显著。时序视角和交互视角的结合充分利用了各自的优势，实现了更全面的特征提取和表征。通过CNN和LSTM捕捉负载和时序特征，再通过GCN分析流量交互图的结构特征，最终融合这些多视角特征，显著提升了异常流量的检测精度和鲁棒性。

消融实验结果表明，时序特征和交互特征的融合显著提升了模型的整体性能，验证了本文所提出的多视角特征融合方法在网络异常流量检测中的有效性和优越性。

## 5　参数分析

### 5.1　数据包长度与负载字节数

在本文所提出的方法中，使用一个流中的前$n$个数据包作为一条网络流量的输入，使用$m$个字节作为单个数据包负载表示。由于从流量中提取的数据包个数以及负载字节数会直接影响模型的训练效率及其性能，对这两个参数进行了实验分析。根据文献[36]分析，$n$和$m$的取值过大会导致模型效率大幅下降，因此，本文实验设置的数据包数量范围从10到100，负载字节数范围从16到32，对多种组合参数进行实验，其实验结果如图4所示。

实验结果显示，在数据包数量$n$从10不断增长到40的过程中，模型效果在稳步提升。而在数据包数量达到40以上后，模型的检测性能并不能得到进一步的提升，因为一般的流量中，数据包个数大多在1~40之间，在使用更多的数据包过程中，由于序列数据的特殊性，需要在数量不足$n$的网络流量中填充空白包，这会导致每个流中存在过多的无意义数据，影响实验精度。而在对负载字节数$m$的测试

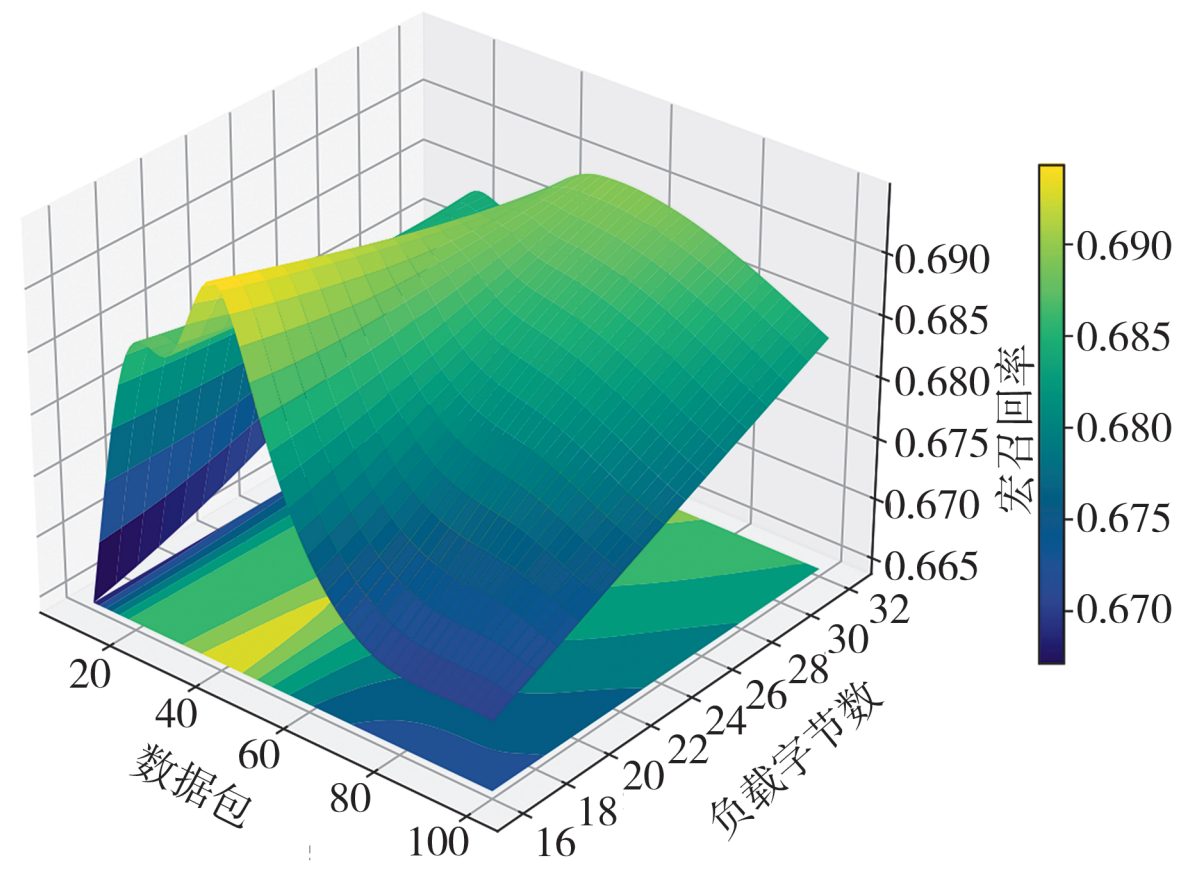

图4　流数据包数量$n$与负载字节数$m$的组合评估

Fig. 4　Combined evaluation of the number of flow packets $n$ and the payload byte count $m$

中可见，在相同数据包数量$n$的情况下，更短的负载字节信息更有利于模型对于不同类型流量的检测，因为在网络数据包的传输过程中，并非所有的数据包都具有负载报文信息。而针对序列数据的填充操作，可能会导致模型获得的无意义信息过多，从而导致性能下降。因此，综合模型效率和性能进行考虑，选择参数（$n$=40，$m$=16）作为最终的实验参数。

### 5.2　视角融合权重参数

在特征融合阶段，利用一个权重系数$\alpha$来对网络流量的时序特征和交互特征进行融合。为了进一步分析不同视角特征对异常流量检测的贡献，对$\alpha$进行了参数分析，结果如图5所示。可见，当权重系数被设定为$\alpha = 0.5$时，模型在大多数情况下可以取得最优的性能。这表明流量数据的时序特征和交互特征对恶意流量检测有近似的贡献，进一步凸显了多视角特征融合框架的有效性。

## 6　结论

本文创新地结合了时序视角与交互视角对网络流量数据进行分析，并提出了一种多视角特征融合方法来进行恶意流量检测。在6个公开的数据集上进行了大量的实验，验证了所提的方法在识别恶意流量方面的有效性和优越性。同时，通过消融实验，证明了所提出的多视角特征融合方法明显优于单一视角下的检测，具有较好的实际意义与应用价值。未来的工作中，计划进一步拓展和深化MuFF框架的潜力和适用性，考虑融合除时间序列和空间交互数据外的其他异构数据源，持续推动网络流量异常检测技术的进步。

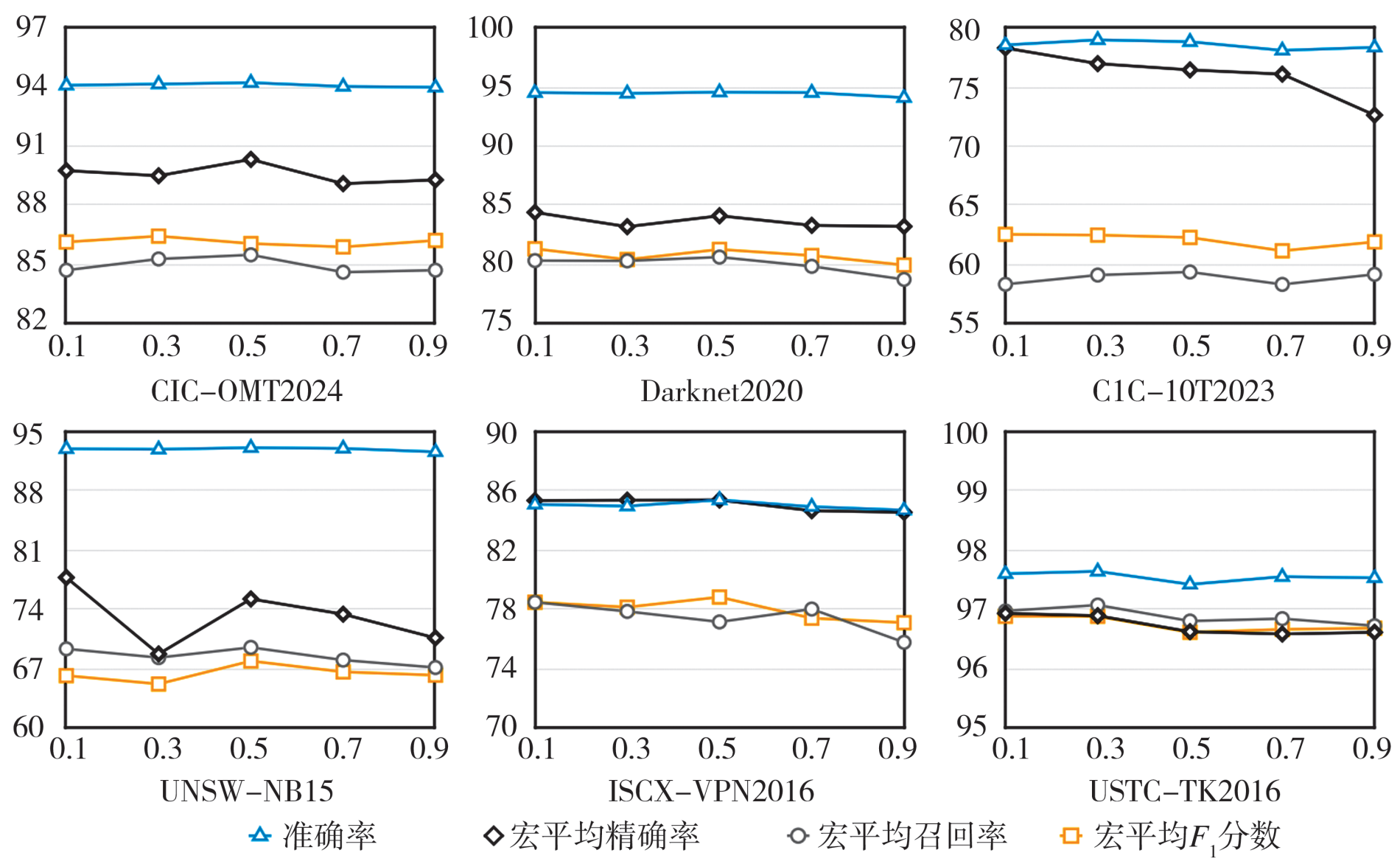

图5　通道贡献度$\alpha$参数评估

Fig. 5　Channel contribution $\alpha$ parameter evaluation

## References

[1] ZHANG J, CHEN X, XIANG Y, et al. Robust network traffic classification[J]. IEEE/ACM Transactions on Networking, 2014, 23(4): 1257−1270.

[2] WEI G, WANG Z. Adoption and realization of deep learning in network traffic anomaly detection device design[J]. Soft Computing, 2021, 25(2): 1147−1158.

[3] KOTPALLIWAR M V, WAJGI R. Classification of attacks using support vector machine (svm) on kddcup'99 idsdatabase[C]// Fifth International Conference on Communication Systems and Network Technologies. IEEE, 2015: 987−990.

[4] KOKILA R, SELVI S T, GOVINDARAJAN K. Ddos detection and analysis in sdn−based environment using support vector machine classifier[C]// Sixth Intenational Conference on Advanced Computing. IEEE, 2014: 205−210.

[5] NGUYEN T T, ARMITAGE G. A survey of techniques for internet traffic classification using machine leaing[J]. IEEE Communications Surveys & Tutorials, 2008, 10(4): 56−76.

[6] KWON D, KIM H, KIM J, et al. A survey of deep leaing based network anomaly detection[J]. Cluster Computing, 2019, 22: 949−961.

[7] LIU H, LANG B. Machine learning and deep learning methods for intrusion detection systems: a survey[J]. Applied Sciences, 2019, 9(20): 4396.

[8] LIN X, XIONG G, GOU G, et al. Et−bert: a contextualized datagram representation with pre−training transformers for encrypted traffic classification[C]// Proceedings of the ACM Web Conference, Lyon, 2022: 633−642.

[9] AZAB A, KHASAWNEH M, ALRABAEE S, et al. Network traffic classification: techniques, datasets, and challenges[J]. Digital Communications and Networks, 2022, 101: 108024.

[10] IZADI S, AHMADI M, NIKBAZM R. Network traffic classification using convolutional neural network and antlion optimization[J]. Computers and Electrical Engineering, 2022, 101: 108024.

[11] WANG Z. The applications of deep learning on traffic identification[J]. BlackHat USA, 2015, 24(11): 1−10.

[12] LIU C, HE L, XIONG G, et al. Fs−net: a flow sequence network for encrypted traffic classification[C]// IEEE INFOCOM IEEE Conference on Computer Communications. IEEE, 2019: 1171−1179.

[13] ZHENG W, ZHONG J, ZHANG Q, et al. Mtt: an efficient model for encrypted network traffic classification using multi−task transformer[J]. Applied Intelligence, 2022, 52(9): 10741−10756.

[14] WANG T, XIE X, WANG W, et al. Netmamba: efficient network traffic classification via pre−training unidirectional mamba[C]// IEEE 32nd International Conference on Network Protocols. IEEE, 2024: 1−11.

[15] WU Z, PAN S, CHEN F, et al. A comprehensive survey on graph neural networks[J]. IEEE Transactions on Neural Net-

works and Learning Systems, 2020, 32(1): 4-24.

[16] HUOH T L, LUO Y, LI P, et al. Flow-based encrypted network traffic classification with graph neural networks[J]. IEEE Transactions on Network and Service Management, 2022, 20(2): 1224-1237.

[17] BARSELLOTTI L, DE MARINIS L, CUGINI F, et al. Ftg-net: hierarchical flow-to-traffic graph neural network for ddos attack detection[C]// IEEE 24th International Conference on High Performance Switching and Routing. IEEE, 2023: 173-178.

[18] HAN X, XU G, ZHANG M, et al. DE-GNN: dual embedding with graph neural network for fine-grained encrypted traffic classification[J]. Computer Networks, 2024, 245: 110372.

[19] YANG Z, MA Z, ZHAO W, et al. Hrnn: hypergraph recurrent neural network for network intrusion detection[J]. Journal of Grid Computing, 2024, 22(2): 1-15.

[20] 赵文博, 马紫彤, 杨哲. 基于超图神经网络的恶意流量分类模型 [J]. 网络与信息安全学报, 2023, 9(5): 166-177.
ZHAO W B, MA Z T, YANG Z. Model of the malicious traffic classification based on hypergraph neural network[J]. Chinese Journal of Network and Information Security, 2023, 9(5): 166-177.(in Chinese)

[21] SHEN M, ZHANG J, ZHU L, et al. Accurate decentralized application identification via encrypted traffic analysis using graph neural networks[J]. IEEE Transactions on Information Forensics and Security, 2021, 16: 2367-2380.

[22] KIPF T N, WELLING M. Semi-supervised classification with graph convolutional networks[C]// International Conference on Learning Representations, 2017: 1-14.

[23] MOUSTAFA N, SLAY J. Unsw-nb15: a comprehensive data set for network intrusion detection systems (unsw-nb15 network data set)[C]// Military Communications and Information Systems Conference. IEEE, 2015: 1-6.

[24] MOUSTAFA N, SLAY J. The evaluation of network anomaly detection systems: statistical analysis of the unsw-nb15 data set and the comparison with the kdd99 data set[J]. Information Security Journal: A Global Perspective, 2016, 25(3): 18-31.

[25] MOUSTAFA N, SLAY J, CREECH G. Novel geometric area analysis technique for anomaly detection using trapezoidal area estimation on large-scale networks[J]. IEEE Transactions on Big Data, 2017, 5(4): 481-494.

[26] SARHAN M, LAYEGHY S, MOUSTAFA N, et al. Net-flow datasets for machine learning-based network intrusion detection systems[C]// Big Data Technologies and Applications: 10th EAI International Conference, BDTA 2020, and 13th EAI International Conference on Wireless Internet, Springer, 2021: 117-135.

[27] HABIBI LASHKARI A, KAUR G, RAHALI A. Didarknet: a contemporary approach to detect and characterize the darknet traffic using deep image learning[C]// Proceedings of the 2020 10th International Conference on Communication and Network Security, 2020: 1-13.

[28] DRAPER-GIL G, LASHKARI A H, MAMUN M S I, et al. Characterization of encrypted and vpn traffic using time-related[C]// Proceedings of the 2nd International Conference on Information Systems Security and Privacy, 2016: 407-414.

[29] NETO E C P, DADKHAH S, FERREIRA R, et al. Ciciot2023: a real-time dataset and benchmark for large-scale attacks in iot environment[J]. Sensors, 2023, 23(13): 5941.

[30] WANG W, ZHU M, ZENG X, et al. Malware traffic classification using convolutional neural network for representation learning[C]// International Conference on Information Networking. IEEE, 2017: 712-717.

[31] HAMILTON W, YING Z, LESKOVEC J. Inductive representation learning on large graphs[J]. Advances in Neural Information Processing Systems, 2017, 30: 1-11.

[32] XU K, HU W, LESKOVEC J, et al. How powerful are graph neural networks[C]// International Conference on Learning Representations, 2019: 1-17.

[33] VELIČKOVIĆ P, CUCURULL G, CASANOVA A, et al. Graph attention networks[J]. Statistics, 2017, 20: 1048850.

[34] LO W W, LAYEGHY S, SARHAN M, et al. E-graphsage: a graph neural network based intrusion detection system for iot [C]// NOMS IEEE/IFIP Network Operations and Management Symposium. IEEE, 2022: 1-9.

[35] LI Y, LI R, ZHOU Z, et al. Graphddos: effective ddos attack detection using graph neural networks[C]// IEEE 25th International Conference on Computer Supported Cooperative Work in Design. IEEE, 2022: 1275-1280.

[36] BIKMUKHAMEDOV R, NADEEV A. Multi-class net-work traffic generators and classifiers based on neural networks [C]// Systems of Signals Generating and Processing in the Field of on Board Communications. IEEE, 2021: 1-7.